\documentclass[11pt]{article}
\usepackage{acl2016}
\usepackage{times}
\usepackage{latexsym}
\usepackage{amsmath}
\usepackage{bm}
\usepackage{algorithm}
\usepackage{algorithmic}
\usepackage{amsmath,blkarray,bbold,graphicx}
\usepackage[colorinlistoftodos]{todonotes}
\usepackage{url}
\usepackage{amsmath}
\usepackage{multirow}
\usepackage{breqn}

\newcommand\ie {{\it i.e., }}
\newcommand\eg {{\it e.g., }}

\aclfinalcopy 
\def\aclpaperid{***} 

\title{Nonparametric Spherical Topic Modeling with Word Embeddings}

\author{Kayhan Batmanghelich\thanks{ \hspace{1.5mm}Authors contributed equally and listed alphabetically.}\\
	    CSAIL, MIT\\
	    {\tt kayhan@mit.edu}\\
	  \And
	Ardavan Saeedi\textsuperscript{\scriptsize{*}}\\
  	CSAIL, MIT\\
  {\tt ardavans@mit.edu}
  \And 
  Karthik Narasimhan \\ 
  CSAIL, MIT\\
  {\tt karthikn@mit.edu}
  \AND 
  Sam Gershman\\
	    Harvard University\\
	    {\tt gershman@fas.harvard.edu}}

\date{}

\begin{document}

\maketitle
\vspace{2in}
\begin{abstract}
Traditional topic models do not account for semantic regularities in language. 
Recent distributional representations of words exhibit semantic consistency over directional metrics such as cosine similarity. 
However, neither categorical nor Gaussian observational distributions used in existing topic models are appropriate to leverage such correlations.
In this paper, we propose to use the von Mises-Fisher distribution to model the density of words over a unit sphere. Such a representation is well-suited for directional data. We use a Hierarchical Dirichlet Process for our base topic model and propose an efficient inference algorithm based on Stochastic Variational Inference. This model enables us to naturally exploit the semantic structures of word embeddings while flexibly discovering the number of topics. Experiments demonstrate that our method outperforms competitive approaches in terms of topic coherence on two different text corpora while offering efficient inference.
\end{abstract}

\section{Introduction}
Prior work on topic modeling has mostly involved the use of categorical likelihoods~\cite{blei2003latent,blei2006dynamic,rosen2004author}. Applications of topic models in the textual domain treat words as discrete observations, ignoring the semantics of the language. Recent developments in distributional representations of words~\cite{mikolov2013distributed,pennington2014glove} have succeeded in capturing certain semantic regularities, but have not been explored extensively in the context of topic modeling. In this paper, we propose a probabilistic topic model with a novel observational distribution that integrates well with directional similarity metrics.


One way to employ semantic similarity is to use the Euclidean distance between word vectors, which reduces to a Gaussian observational distribution for topic modeling~\cite{das2015gaussian}. 
The \emph{cosine distance} between word embeddings is another popular choice and has been shown to be a good measure of semantic relatedness~\cite{mikolov2013distributed,pennington2014glove}.
The von Mises-Fisher (vMF) distribution is well-suited to model such directional data~\cite{dhillon2003modeling,banerjee2005clustering} but has not been previously applied to topic models. 

In this work, we use vMF as the observational distribution. Each word can be viewed as a point on a unit sphere with topics being canonical directions. More specifically, we use a Hierarchical Dirichlet Process (HDP)~\cite{teh2006hierarchical}, a Bayesian nonparametric variant of Latent Dirichlet Allocation (LDA), to automatically infer the number of topics. We implement an efficient inference scheme based on Stochastic Variational Inference (SVI)~\cite{hoffman2013stochastic}.

We perform experiments on two different English text corpora: \textsc{20 Newsgroups} and \textsc{Nips} and compare against two baselines - HDP and Gaussian LDA.
Our model, spherical HDP (sHDP), outperforms all three systems on the measure of \emph{topic coherence}. For instance, sHDP obtains gains over Gaussian LDA of 97.5\% on the \textsc{Nips} dataset and 65.5\% on the \textsc{20 Newsgroups} dataset. Qualitative inspection reveals consistent topics produced by sHDP. We also empirically demonstrate that employing SVI leads to efficient topic inference.


\section{Related Work}
\paragraph{Topic modeling and word embeddings}
\newcite{das2015gaussian} proposed a topic model which uses a Gaussian distribution over word embeddings.
By performing inference over the vector representations of the words, their model is encouraged to group words that are semantically similar, leading to more coherent topics. In contrast, we propose to utilize von Mises-Fisher (vMF) distributions which rely on the cosine similarity between the word vectors instead of euclidean distance.

\paragraph{vMF in topic models}
The vMF distribution has been used to model directional data by placing points on a unit sphere~\cite{dhillon2003modeling}. \newcite{reisinger2010spherical} propose an admixture model that uses vMF to model documents represented as vector of normalized word frequencies. This does not account for word level semantic similarities. Unlike their method, we use vMF over word embeddings. In addition, our model is nonparametric. 

\paragraph{Nonparametric topic models}
HDP and its variants have been successfully applied to topic modeling~ \cite{paisley2015nested,blei2012probabilistic,he2013dynamic}; however, all these models assume a categorical likelihood in which the words are encoded as one-hot representation.

\section{Model}
In this section, we describe the generative process for documents. Rather than one-hot representation of words, we employ normalized word embeddings \cite{mikolov2013distributed} to capture semantic meanings of associated words. Word $n$ from document $d$ is represented by a normalized $M$-dimensional vector $x_{dn}$ and the similarity between words is quantified by the cosine of angle between the corresponding word vectors. 

\begin{figure}[bt]
\begin{center}
\includegraphics[width=0.5\columnwidth ]{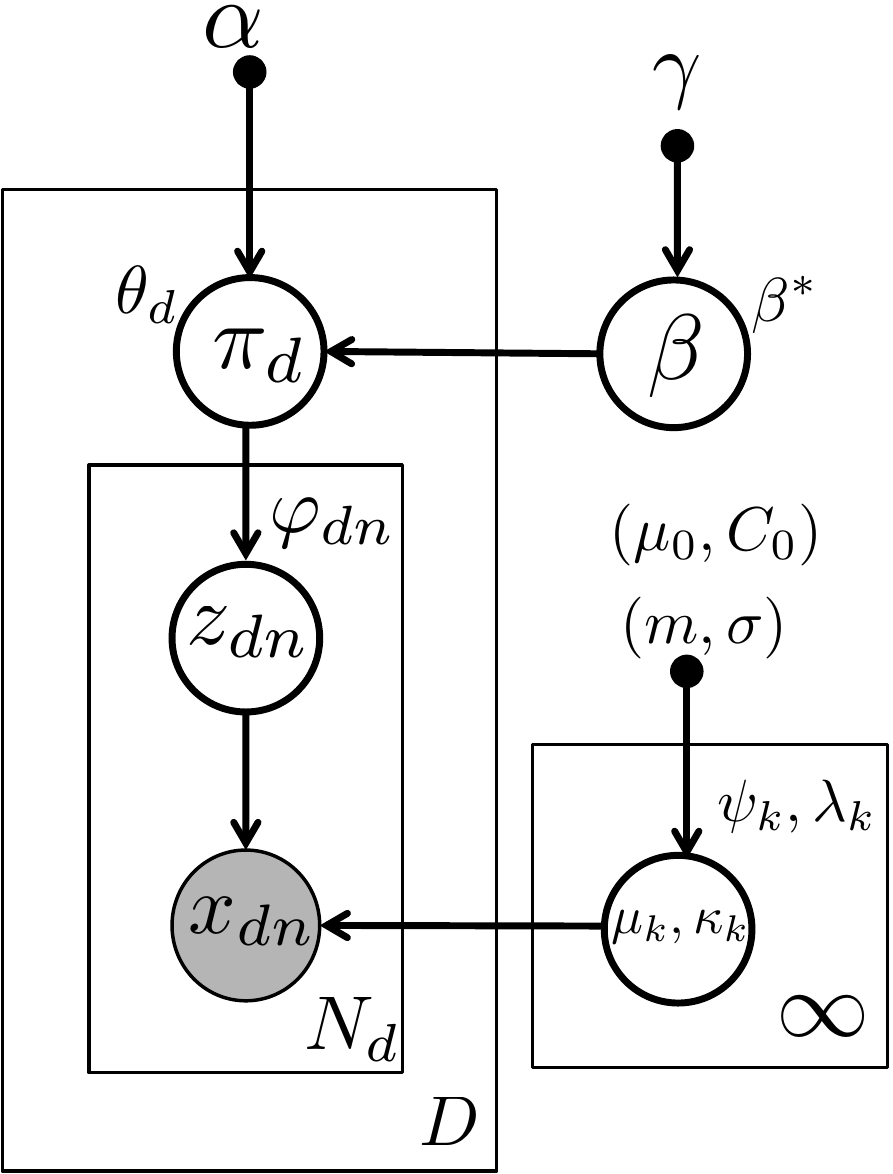} 
\caption{{Graphical representation of our spherical HDP (sHDP) model. The symbol next to each random variable denotes the parameter of its variational distribution. We assume $D$ documents in the corpus, each document contains $N_d$ words and there are countably infinite topics represented by $(\mu_k, \kappa_k)$}.}
\label{fig:graphicalmodel}
\end{center}
\end{figure}

Our model is based on the Hierarchical Dirichlet Process (HDP). The model assumes a collection of ``topics'' that are shared across documents in the corpus. The topics are represented by the topic centers $\mu_k \in \mathbb{R}^M$. Since word vectors are normalized, the $\mu_k$ can be viewed as a direction on unit sphere. 
Von Mises$-$Fisher (vMF) is a distribution that is commonly used to model directional data.  The likelihood of the topic $k$ for word $x_{dn}$ is: 
\begin{equation}
\nonumber
     f(x_{dn}; \mu_k; \kappa_k ) = \exp \left( \kappa_k \mu_k^T x_{dn} \right)  C_{M}(\kappa_k) 
\end{equation}    
where $\kappa_k$ is the concentration of the topic $k$, the $C_{M}(\kappa_k):= \kappa_k^{M/2 - 1} /  \left( (2 \pi)^{M/2}  I_{M/2 - 1}(\kappa_k) \right)$ is the normalization constant, and $I_{\nu}(\cdot)$ is the modified Bessel function of the first kind at order $\nu$. Interestingly, the log-likelihood of the vMF is proportional to  $\mu_k^T x_{dn}$ (up to a constant), which is equal to the cosine distance between two vectors. This distance metric is also used in \newcite{mikolov2013distributed} to measure semantic proximity.

When sampling a new document, a subset of topics determine the distribution over words. We let $z_{dn}$ denote the topic selected for the word $n$ of document $d$. Hence, $z_{dn}$ is drawn from a categorical distribution: $z_{dn} \sim \text{Mult}(\pi_d)$, where $\pi_d$ is the proportion of topics for document $d$. We draw $\pi_{d}$ from a Dirichlet Process which enables us to estimate the the number of topics from the data. The generative process for the generation of new document is as follows:
\begin{align*}
\nonumber
&\beta \sim \text{GEM}(\gamma) &&\quad \pi_d \sim \text{DP}(\alpha,\beta)\\
&\kappa_k \sim \text{log-Normal}(m, \sigma^2) && \quad \mu_k \sim \text{vMF}(\mu_0, C_0)  \\
&z_{dn} \sim \text{Mult}(\pi_d) &&\quad x_{dn} \sim \text{vMF}(\mu_k, \kappa_k)
\end{align*}
where GEM$(\gamma)$ is the stick-breaking distribution with concentration parameter $\gamma$, DP$(\alpha,\beta)$ is a Dirichlet process with concentration parameter $\alpha$ and stick proportions $\beta$ \cite{teh2012hierarchical}. We use log-normal and vMF as hyper-prior distributions for the concentrations ($\kappa_k$) and centers of the topics ($\mu_k$) respectively. Figure~\ref{fig:graphicalmodel} provides a graphical illustration of the model.

\paragraph{Stochastic variational inference}
\label{sec:svi}
In the rest of the paper, we use bold symbols to denote the variables of the same kind (\eg~$\bm{x}_d = \{ x_{dn} \}_n$, $\bm{z}:=\{ z_{dn} \}_{d,n}$). 
We employ stochastic variational mean-field inference (SVI)~\cite{hoffman2013stochastic} to estimate the posterior distributions of the latent variables.  SVI enables us to sequentially process  batches of documents  which makes it appropriate in large-scale settings. 

To approximate the posterior distribution of the latent variables, the mean-field approach finds the optimal parameters of the fully factorizable $q$ (\ie $q(\textbf{z}, \beta, \bm{\pi}, \bm{\mu}, \bm{\kappa}):=q(\textbf{z})q(\beta)q(\bm{\pi})q(\bm{\mu}) q(\bm{\kappa})$) by maximizing the Evidence Lower Bound (ELBO),
\begin{equation}
\nonumber
\mathcal{L}(q) = \mathbb{E}_q \left[ \log p(\bm{X}, \textbf{z}, \beta, \bm{\pi}, \bm{\mu}, \bm{\kappa})  \right] - \mathbb{E}_q \left[ \log q \right]
\end{equation}
where $\mathbb{E}_q[\cdot]$ is expectation with respect to $q$, $p(\bm{X}, \textbf{z}, \beta, \bm{\pi}, \bm{\mu}, \bm{\kappa})$ is the joint likelihood of the model specified by the HDP model. 


The variational distributions for $\bm{z},\bm{\pi},\bm{\mu}$ have the following parametric forms,
\begin{align*}
\nonumber
q(\textbf{z}) &= \text{Mult}(\mathbf{z}|\bm{\varphi}) \\
q(\bm{\pi}) &= \text{Dir}(\bm \pi|\bm{\theta})\\
q(\bm{\mu}) &= \text{vMF}(\bm \mu|\bm{\psi},\bm{\lambda}),
\end{align*}
where $\text{Dir}$ denotes the Dirichlet distribution and $\bm{\varphi},\bm{\theta},\bm{\psi}$ and $\bm{\lambda}$ are the parameters we need to optimize the ELBO. Similar to \cite{bryant2012truly}, we view $\beta$ as a parameter; hence, $q(\beta) = \delta_{\beta^{*}}(\beta)$. The prior distribution $\kappa$ does not follow a conjugate distribution; hence, its posterior does not have a closed-form. Since $\kappa$ is only one dimensional variable, we use importance sampling to approximate its posterior.  
For a batch size of one (\ie processing one document at time), the update equations for the parameters are: 
\begin{equation}
\begin{split}
\nonumber
&\varphi_{dwk} \propto \exp\{\mathbb{E}_q[\log \text{vMF}(x_{dw}|\psi_k,\lambda_k)] \\ & \qquad \qquad  +\mathbb{E}_q[\log\pi_{dk}]\}\\
&{\theta}_{dk} \leftarrow (1-\rho) {\theta}_{dk} + \rho (\alpha \beta_{k} + D\sum_{n=1}^{W}\omega_{wj}\varphi_{dwk})\\
&t  \leftarrow (1 - \rho) t + \rho  s(\bm{x}_d , \varphi_{dk} ) \\
& \psi \leftarrow  t/\| t \|_2, \quad \lambda \leftarrow \| t \|_2
\end{split}
\end{equation}
where $D$, $\omega_{wj}$, $W$, $\rho$  are  the total number of documents,  number of word $w$ in document $j$, the total number of words in the dictionary, and the step size, respectively. $t$ is a natural parameter for vMF and $s(\bm{x}_d , \varphi_{dk})$ is a function computing the sufficient statistics of vMF distribution of the topic $k$. 
We use numerical gradient ascent to optimize for $\beta^{*}$ (see \newcite{gopal2014mises} for exact forms of $\mathbb{E}_q \log [ \text{vMF}(x_{dw}|\psi_k,\lambda_k) ]$ and $\mathbb{E}_q[\log\pi_{dk}]$ ).

\section{Experiments}

\paragraph{Setup}
We perform experiments on two different text corpora: 11266 documents from \textsc{20 Newsgroups}\footnote{\url{http://qwone.com/~jason/20Newsgroups/}} and 1566 documents from the \textsc{Nips} corpus\footnote{\url{http://www.cs.nyu.edu/~roweis/data.
html}}. We utilize 50-dimensional word embeddings trained on text from Wikipedia using \emph{word2vec}\footnote{https://code.google.com/p/word2vec/
}. The vectors are post-processed to have unit $\ell^2$-norm. We evaluate our model using the measure of topic coherence~\cite{newman2010automatic}, which has been shown to effectively correlate with human judgement~\cite{lau2014machine}. For this, we compute the Pointwise Mutual Information (PMI) using a reference corpus of 300k documents from Wikipedia. The PMI is calculated using co-occurence statistics over pairs of words ($u_i$, $u_j$) in 20-word sliding windows:
\begin{equation*}
\text{PMI}(u_i, u_j) = \log \frac{p(u_i, u_j)}{p(u_i) \cdot p(u_j)}
\end{equation*}
We compare our model with two baselines: HDP and the Gaussian LDA model .
We ran G-LDA with various number of topics ($k$).

\paragraph{Results}

 \begin{table*}[t]
\centering
\resizebox{\textwidth}{!}{%
\begin{tabular}{ c c c c c c c c } \hline
\multicolumn{8}{c}{Gaussian LDA} \\ \hline
vector & shows & network & hidden & performance & net & figure & size  \\
image & feature & learning & term & work & references & shown & average  \\
gaussian & show & model & rule & press & introduction & neurons & present  \\
equation & motion & neural & word & tion & statistical & point & family  \\
generalization & action & input & means & ing & related & large & versus  \\
images & spike & data & words & eq & comparison & neuron & spread  \\
gradient & series & function & approximate & performed & source & small & median  \\
theory & final & time & derived & em & statistics & fig & physiology  \\
dimensional & robot & set & describe & vol & free & cells & children  \\ \hline
1.16 & 0.4 & 0.35 & 0.29 & 0.25 & 0.25 & 0.21 & 0.2 \\ \hline
\multicolumn{8}{c}{Spherical HDP} \\ \hline
neural & function & analysis & press & pattern & problem & noise & algorithm  \\
layer & linear & theory & cambridge & fig & process & gradient & error  \\
neurons & functions & computational & journal & temporal & method & propagation & parameters  \\
neuron & vector & statistical & vol & shape & optimal & signals & computation  \\
activation & random & field & eds & smooth & solution & frequency & algorithms  \\
brain & probability & simulations & trans & surface & complexity & feedback & compute  \\
cells & parameter & simulation & springer & horizontal & estimation & electrical & binary  \\
cell & dimensional & nonlinear & volume & vertical & prediction & filter & mapping  \\
synaptic & equation & dynamics & review & posterior & solve & detection & optimization  \\ \hline
1.87 & 1.73 & 1.51 & 1.44 & 1.41 & 1.19 & 1.12 & 1.03 \\ \hline
\end{tabular}
}
\caption{Examples of top words for the most coherent topics (column-wise) inferred on the \textsc{Nips} dataset by Gaussian LDA (k=40) and Spherical HDP. The last row for each model is the topic coherence (PMI) computed using Wikipedia documents as reference.}
\label{table:qualitative}
\end{table*}

 \begin{table}[t]
\centering
\begin{tabular}{ | c | c   c | } \hline
\multirow{2}{*}{\textbf{Model}} & \multicolumn{2}{c |}{\textbf{Topic Coherence}} \\ 
& \textsc{20 News} & \textsc{Nips} \\ \hline
HDP & 0.037 & 0.270 \\
G-LDA (k=20) & -0.017 & 0.215 \\
G-LDA (k=40) & 0.052 & 0.248 \\
G-LDA (k=60) & 0.082 & 0.137 \\
G-LDA (k=100) & -0.032 & 0.267 \\
sHDP & \textbf{0.162} & \textbf{0.442} \\ \hline
\end{tabular}
\caption{Average topic coherence for various baselines (HDP, Gaussian LDA (G-LDA)) and sHDP. $k$=number of topics. Best scores are shown in bold.}
\label{table:results}
\end{table}

Table~\ref{table:results} details the topic coherence averaged over all topics produced by each model. We observe that our sHDP model outperforms G-LDA by 0.08 points on \textsc{20 Newsgroups} and by 0.17 points on the \textsc{Nips} dataset. We can also see that the individual topics inferred by sHDP make sense qualitatively and have higher coherence scores than G-LDA (Table \ref{table:qualitative}). This supports our hypothesis that using the vMF likelihood helps in producing more coherent topics. sHDP produces 16 topics for the \textsc{20 Newsgroups} and 92 topics on the \textsc{Nips} dataset.

\begin{figure}[bt]
\begin{center}
\includegraphics[width=0.9\columnwidth ]{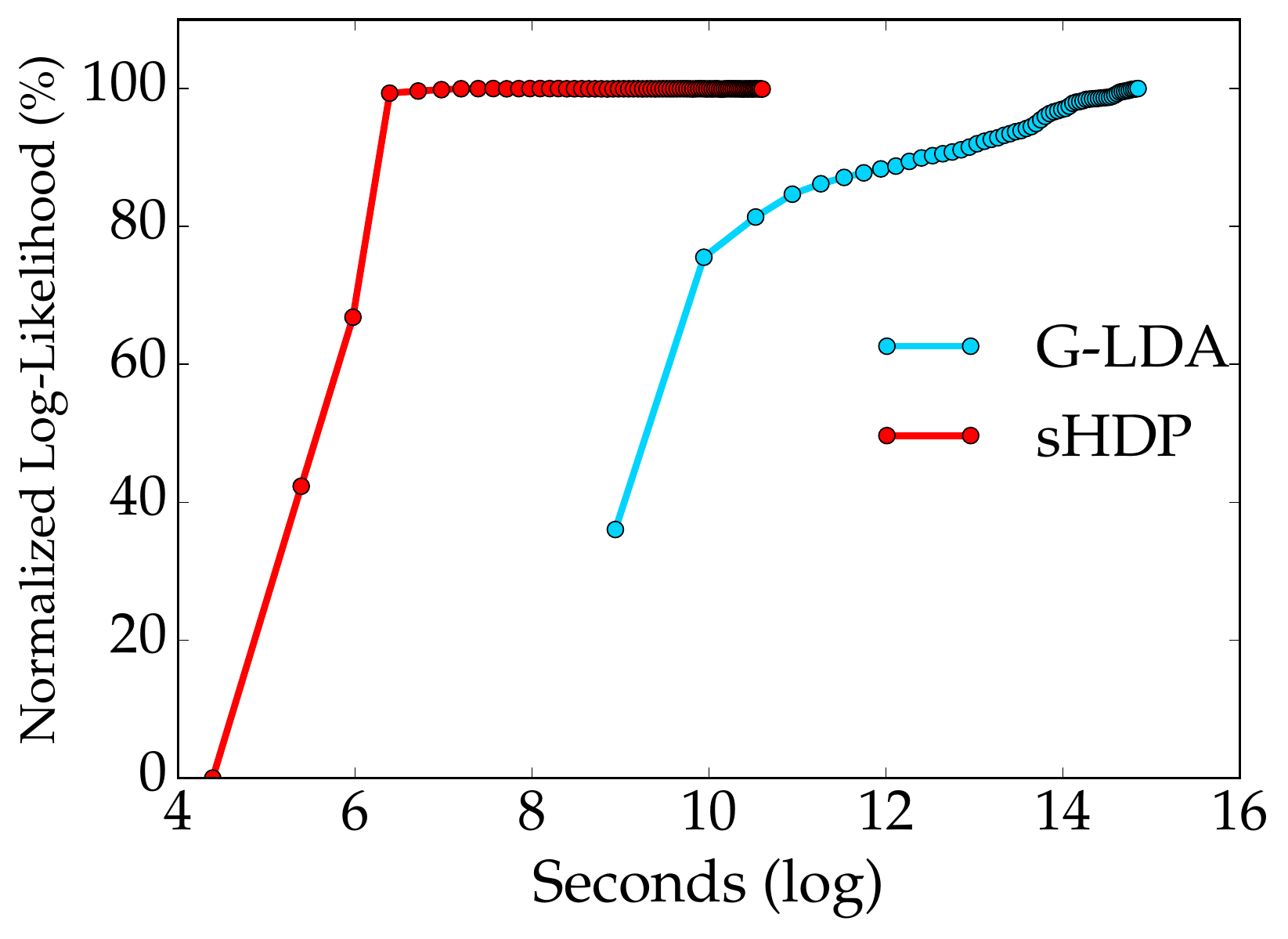} 
\vspace{-3mm}
\caption{{Normalized log-likelihood (in percentage) over a training set of size 1566 documents from the NIPS corpus. Since the log-likelihood values are not comparable for the Gaussian LDA and the sHDP, we normalize them to demonstrate the convergence speed of the two inference schemes for these models.}}
\label{fig:runtime}
\end{center}
\end{figure}

Figure~\ref{fig:runtime} shows a plot of normalized log-likelihood against the runtime of sHDP and G-LDA.\footnote{Our sHDP implementation is in Python and the G-LDA code is in Java.} We calculate the normalized value of log-likelihood by subtracting the minimum value from it and dividing it by the difference of maximum and minimum values. We can see that sHDP converges faster than G-LDA, requiring only around five iterations while G-LDA takes longer to converge.

\section{Conclusion}

Classical topic models do not account for semantic regularities in language. Recently, distributional representations of words have emerged that exhibit semantic consistency over directional metrics like cosine similarity. Neither categorical nor Gaussian observational distributions used in existing topic models are appropriate to leverage such correlations. In this work, we demonstrate the use of the von Mises-Fisher distribution to model words as points over a unit sphere. We use HDP as the base topic model and propose an efficient algorithm based on Stochastic Variational Inference. Our model  naturally exploits the semantic structures of word embeddings while flexibly inferring the number of topics. We show that our method outperforms three competitive approaches in terms of topic coherence on two different datasets.


\bibliography{acl2016}

\begin{thebibliography}{}

\bibitem[\protect\citename{Banerjee \bgroup et al.\egroup
  }2005]{banerjee2005clustering}
Arindam Banerjee, Inderjit~S Dhillon, Joydeep Ghosh, and Suvrit Sra.
\newblock 2005.
\newblock Clustering on the unit hypersphere using von mises-fisher
  distributions.
\newblock In {\em Journal of Machine Learning Research}, pages 1345--1382.

\bibitem[\protect\citename{Blei and Lafferty}2006]{blei2006dynamic}
David~M Blei and John~D Lafferty.
\newblock 2006.
\newblock Dynamic topic models.
\newblock In {\em Proceedings of the 23rd international conference on Machine
  learning}, pages 113--120. ACM.

\bibitem[\protect\citename{Blei \bgroup et al.\egroup }2003]{blei2003latent}
David~M Blei, Andrew~Y Ng, and Michael~I Jordan.
\newblock 2003.
\newblock Latent dirichlet allocation.
\newblock {\em the Journal of machine Learning research}, 3:993--1022.

\bibitem[\protect\citename{Blei}2012]{blei2012probabilistic}
David~M Blei.
\newblock 2012.
\newblock Probabilistic topic models.
\newblock {\em Communications of the ACM}, 55(4):77--84.

\bibitem[\protect\citename{Bryant and Sudderth}2012]{bryant2012truly}
Michael Bryant and Erik~B Sudderth.
\newblock 2012.
\newblock Truly nonparametric online variational inference for hierarchical
  dirichlet processes.
\newblock In {\em Advances in Neural Information Processing Systems}, pages
  2699--2707.

\bibitem[\protect\citename{Das \bgroup et al.\egroup }2015]{das2015gaussian}
Rajarshi Das, Manzil Zaheer, and Chris Dyer.
\newblock 2015.
\newblock Gaussian {LDA} for topic models with word embeddings.
\newblock In {\em Proceedings of the 53nd Annual Meeting of the Association for
  Computational Linguistics}.

\bibitem[\protect\citename{Dhillon and Sra}2003]{dhillon2003modeling}
Inderjit~S Dhillon and Suvrit Sra.
\newblock 2003.
\newblock Modeling data using directional distributions.
\newblock Technical report, Technical Report TR-03-06, Department of Computer
  Sciences, The University of Texas at Austin. URL ftp://ftp. cs. utexas.
  edu/pub/techreports/tr03-06. ps. gz.

\bibitem[\protect\citename{Gopal and Yang}2014]{gopal2014mises}
Siddarth Gopal and Yiming Yang.
\newblock 2014.
\newblock Von mises-fisher clustering models.

\bibitem[\protect\citename{He \bgroup et al.\egroup }2013]{he2013dynamic}
Yulan He, Chenghua Lin, Wei Gao, and Kam-Fai Wong.
\newblock 2013.
\newblock Dynamic joint sentiment-topic model.
\newblock {\em ACM Transactions on Intelligent Systems and Technology (TIST)},
  5(1):6.

\bibitem[\protect\citename{Hoffman \bgroup et al.\egroup
  }2013]{hoffman2013stochastic}
Matthew~D Hoffman, David~M Blei, Chong Wang, and John Paisley.
\newblock 2013.
\newblock Stochastic variational inference.
\newblock {\em The Journal of Machine Learning Research}, 14(1):1303--1347.

\bibitem[\protect\citename{Lau \bgroup et al.\egroup }2014]{lau2014machine}
Jey~Han Lau, David Newman, and Timothy Baldwin.
\newblock 2014.
\newblock Machine reading tea leaves: Automatically evaluating topic coherence
  and topic model quality.
\newblock In {\em EACL}, pages 530--539.

\bibitem[\protect\citename{Mikolov \bgroup et al.\egroup
  }2013]{mikolov2013distributed}
Tomas Mikolov, Ilya Sutskever, Kai Chen, Greg~S Corrado, and Jeff Dean.
\newblock 2013.
\newblock Distributed representations of words and phrases and their
  compositionality.
\newblock In {\em Advances in neural information processing systems}, pages
  3111--3119.

\bibitem[\protect\citename{Newman \bgroup et al.\egroup
  }2010]{newman2010automatic}
David Newman, Jey~Han Lau, Karl Grieser, and Timothy Baldwin.
\newblock 2010.
\newblock Automatic evaluation of topic coherence.
\newblock In {\em Human Language Technologies: The 2010 Annual Conference of
  the North American Chapter of the Association for Computational Linguistics},
  pages 100--108. Association for Computational Linguistics.

\bibitem[\protect\citename{Paisley \bgroup et al.\egroup
  }2015]{paisley2015nested}
John Paisley, Chingyue Wang, David~M Blei, and Michael~I Jordan.
\newblock 2015.
\newblock Nested hierarchical dirichlet processes.
\newblock {\em Pattern Analysis and Machine Intelligence, IEEE Transactions
  on}, 37(2):256--270.

\bibitem[\protect\citename{Pennington \bgroup et al.\egroup
  }2014]{pennington2014glove}
Jeffrey Pennington, Richard Socher, and Christopher~D. Manning.
\newblock 2014.
\newblock Glove: Global vectors for word representation.
\newblock In {\em Empirical Methods in Natural Language Processing (EMNLP)},
  pages 1532--1543.

\bibitem[\protect\citename{Reisinger \bgroup et al.\egroup
  }2010]{reisinger2010spherical}
Joseph Reisinger, Austin Waters, Bryan Silverthorn, and Raymond~J Mooney.
\newblock 2010.
\newblock Spherical topic models.
\newblock In {\em Proceedings of the 27th International Conference on Machine
  Learning (ICML-10)}, pages 903--910.

\bibitem[\protect\citename{Rosen-Zvi \bgroup et al.\egroup
  }2004]{rosen2004author}
Michal Rosen-Zvi, Thomas Griffiths, Mark Steyvers, and Padhraic Smyth.
\newblock 2004.
\newblock The author-topic model for authors and documents.
\newblock In {\em Proceedings of the 20th conference on Uncertainty in
  artificial intelligence}, pages 487--494. AUAI Press.

\bibitem[\protect\citename{Teh \bgroup et al.\egroup
  }2006]{teh2006hierarchical}
Yee~Whye Teh, Michael~I Jordan, Matthew~J Beal, and David~M Blei.
\newblock 2006.
\newblock Hierarchical dirichlet processes.
\newblock {\em Journal of the American Statistical Association},
  101:1566--1581.

\bibitem[\protect\citename{Teh \bgroup et al.\egroup
  }2012]{teh2012hierarchical}
Yee~Whye Teh, Michael~I Jordan, Matthew~J Beal, and David~M Blei.
\newblock 2012.
\newblock Hierarchical dirichlet processes.
\newblock {\em Journal of the american statistical association}.

\end{thebibliography}
\bibliographystyle{acl2016}

\end{document}